\pdfoutput=1

\documentclass[11pt]{article}

\usepackage[preprint]{acl}

\usepackage{times}
\usepackage{latexsym}

\usepackage[T1]{fontenc}

\usepackage[utf8]{inputenc}

\usepackage{microtype}

\usepackage{inconsolata}

\usepackage{graphicx}
\usepackage{multirow}
\usepackage{pifont}
\usepackage{xcolor}
\usepackage{booktabs}
\usepackage{hyperref}
\usepackage{makecell}
\usepackage{amsmath}
\usepackage[tikz]{bclogo}
\usepackage{amsfonts}
\usepackage{enumitem}
\usepackage{stackengine}
\usepackage{multicol}
\usepackage{bm}
\usepackage[ruled,vlined]{algorithm2e}
\usepackage{algpseudocode}
\usepackage{soul}
\usepackage{colortbl}
\usepackage{enumitem}

\usepackage[most]{tcolorbox}
\tcbuselibrary{skins}
\tcbuselibrary{breakable}

\definecolor{my_blue}{RGB}{0,120,255}
\definecolor{my_purple}{RGB}{161, 27, 155}
\definecolor{my_green}{RGB}{0, 176, 80}
\definecolor{msftBlue}{RGB}{0,164,239}
\definecolor{msftGreen}{RGB}{127,186,0}
\definecolor{msftYello}{RGB}{255,185,0}
\definecolor{msftBlack}{RGB}{0,0,0}
\newcommand{\finding}[1]{
\begin{bclogo}[couleur= msftBlack!05, epBord= 1, arrondi=0.1, logo=\bclampe,marge= 2, ombre=true, blur, couleurBord=msftBlack!10, tailleOndu=3, sousTitre ={\em #1}]{} 
\end{bclogo}
}

\definecolor{my-blue}{RGB}{0, 69, 177}
\definecolor{my-purple}{RGB}{105,50,230}

\title{More Tokens, Lower Precision: Towards the Optimal Token-Precision \\Trade-off in KV Cache Compression}

\author{%
  Jiebin Zhang~$^\spadesuit$ \quad Dawei Zhu~$^\spadesuit$ \quad Yifan Song~$^\spadesuit$ \quad Wenhao Wu~$^\spadesuit$ \\ \textbf{Chuqiao Kuang~$^\clubsuit$ \quad  Xiaoguang Li~$^\clubsuit$ \quad Lifeng Shang~$^\clubsuit$ \quad Qun Liu~$^\clubsuit$ \quad Sujian Li~$^\spadesuit$ } \\[5pt]
  $^\spadesuit$~Peking University\quad
  $^\clubsuit$~Huawei Technologies \\[5pt]
{{\texttt{zhangjiebin@pku.edu.cn}}}
}

\begin{document}
\maketitle
\begin{abstract}

\renewcommand{\paragraph}[1]{\noindent \textbf{#1} }

As large language models (LLMs) process increasing context windows, the memory usage of KV cache has become a critical bottleneck during inference. The mainstream KV compression methods, including KV pruning and KV quantization, primarily focus on either token or precision dimension separately. However, these works leaving the trade-off between these two orthogonal dimensions largely under-explored. 
In this paper, we comprehensively investigate the token-precision trade-off in KV cache compression. 
Experiments demonstrate that storing more tokens 
in the KV cache with lower precision, 
 a strategy we term \textbf{quantized pruning}, can significantly enhance the long-context performance of LLMs. 
In-depth analysis of the token-precision trade-off across key aspects demonstrates that, quantized pruning achieves substantial improvements in retrieval-related tasks and consistently performs well across varying input lengths.  
Furthermore,
quantized pruning demonstrates notable stability and effectiveness across different KV pruning methods, quantization strategies, and model scales. These findings offer valuable insights into optimizing KV cache compression through balanced token-precision trade-off strategies. Our code is available at \href{https://github.com/zhzihao/QPruningKV}{https://github.com/zhzihao/QPruningKV}. 
\end{abstract}

\section{Introduction}
Current long-context Large Language Models (LLMs) heavily depend on the Key-Value (KV) cache mechanism, which stores intermediate keys and values during text generation to avoid redundant computations~\citep{6809348}. 
As the input sequence length increases, the KV cache size expands proportionally, leading to substantial memory overhead~\citep{achiam2023gpt,reid2024gemini}. Take Llama3-8B~\citep{dubey2024llama} as an example, storing the KV cache of a single sequence with 100k tokens necessitates a substantial memory overhead of 20GB — exceeding the model's parameter memory of approximately 14GB. Moreover, since the decoding process is highly dependent on GPU memory bandwidth, the large KV cache also results in a significant increase in decoding time~\citep{fu2024challengesdeployinglongcontexttransformers}. As a result, efficiently  compressing the KV cache has become a critical challenge in advancing LLM development~\citep{MLSYS2023_c4be71ab}.

To optimize KV cache memory usage, two primary approaches, focusing on token and precision dimensions, are widely adopted: KV pruning and KV quantization.
For the token dimension, KV pruning methods reduce the KV cache footprint by discarding less salient tokens, thereby maintaining a fixed cache size~\citep{xiao2023streamingllm,10.5555/3666122.3667628,ren2024efficacyevictionpolicykeyvalue,li2024snapkvllmknowslooking}. 
For the precision dimension, KV quantization methods reduce memory usage by approximating KV cache with lower precisions, like 4-bit or even lower~\citep{sheng2023flexgenhighthroughputgenerativeinference,pmlr-v235-liu24bz,hooper2024kvquant10millioncontext,yang2024tokenleftbehindreliable}.
Given the orthogonal nature of token and precision dimensions in KV cache, 
it is natural to consider integrating KV pruning and KV quantization to balance the number of tokens and their precision for better performance. 
For a fixed cache budget, 
lower precision in KV quantization would allow for the retention of more tokens in KV pruning.
This raises the  important questions: Is such an integration effective? How can we achieve an optimal trade-off between pruning and quantization?
To the best of our knowledge, prior research has not explored this interplay.

To address the aforementioned questions, in this paper, we comprehensively explore the token-precision trade-off in KV cache compression.  Through meticulous experimentation, we for the first time present the feasibility of integrating KV pruning and KV quantization.
Specifically, we uncovered an intriguing finding: \textbf{\textit{storing more tokens in the KV cache with lower precision}}, a strategy we term \textbf{\textit{quantized pruning}}, can significantly enhance the long-context performance of LLMs.
For instance, storing $4\times$ tokens in 4-bit precision outperforms storing $1\times$ tokens in 16-bit precision across various KV cache budget. Notably, in extremely low-resource scenarios, quantized pruning effectively preserves performance, whereas relying solely on KV pruning or quantization often leads to a significant performance collapse.
Furthermore, we conduct in-depth analysis of the token-precision trade-off across critical aspects, including task type, input length, model scaling, quantization strategies, and layer-wise configurations. Extensive experiments reveal that quantized pruning achieves considerable gains in retrieval tasks and maintains robust performance across varying input lengths.  Moreover, quantized pruning demonstrates strong stability across different KV pruning methods, quantization strategies, and model scales. 

In summary, our contributions are as follows:
\begin{itemize}[leftmargin=*, nolistsep]
\setlength{\itemsep}{1mm}
     \item We are the first to comprehensively explore the  integration of KV pruning and quantization and propose the novel strategy of quantized pruning to enhance the long-context capabilities of LLMs.
    \item An important finding is revealed: storing more tokens with lower precision significantly outperforms standalone KV pruning or KV quantization methods under various KV cache budgets, highlighting the importance of balancing token count and precision in KV cache compression.
    \item Extensive experiments have been conducted on exploring token-precision trade-off across various critical aspects, including task types, input lengths, model scales, and quantization strategies, providing valuable insights for optimizing KV cache compression.
\end{itemize}

\section{Related Work}

\paragraph{KV Pruning}

KV pruning compresses the KV cache along the token dimension by selectively discarding less salient tokens to reduce memory usage. Mainstream methods typically identify salient tokens based on attention scores, as seen in~\citep{liu2024scissorhands,10.5555/3666122.3667628,oren2024transformersmultistaternns,li2024snapkvllmknowslooking}. Other methods use alternative factors such as initial tokens~\citep{xiao2023streamingllm}, variance~\citep{ren2024efficacyevictionpolicykeyvalue}, special tokens~\citep{ge2024model} or the L2 norm~\citep{devoto2024simpleeffectivel2normbased} to determine token importance. Recent studies delve deeper into optimizing the allocation of KV cache memory budgets. Some explore KV cache budget allocation strategies across layers~\citep{cai2024pyramidkvdynamickvcache,yang2024pyramidinferpyramidkvcache}, while other studies explore head-level KV cache budget allocation~\citep{feng2024adakvoptimizingkvcache,tang2024razorattention,fu2024not,xiao2024duoattention}.

\paragraph{KV Quantization}

KV quantization compress KV cache from the precision dimension by storing KV cache using a reduced number of bits. FlexGen~\citep{sheng2023flexgenhighthroughputgenerativeinference} utilizes group-wise 4-bit quantization for both key and value cache.  KIVI~\citep{pmlr-v235-liu24bz} applies per-channel quantization on key cache and per-token quantization on value cache. KVQuant~\citep{hooper2024kvquant10millioncontext} and CQ~\citep{zhang2024kvcache1bit} use RoPE-related quantization, while KVQuant also preverses outliers without quantization.
Atom~\citep{zhao2024atom} and SKVQ~\citep{duanmu2024skvq} reorders the outlier channels for fine-grained group quantization with mixed-precision. 
QAQ~\citep{dong2024qaqqualityadaptivequantization} and MiKV~\citep{yang2024tokenleftbehindreliable}, inspired by KV pruning methods, store discarded tokens from KV pruning methods using lower bit precision while retaining salient tokens in full precision. However, their memory usage scales proportionally with context length. In contrast, quantized pruning, fundamentally based on KV pruning, theoretically enables compression to a pre-defined memory budget.

\paragraph{Other KV Compression Methods} Compressing KV cache from other dimensions typically requires modifying the model architecture, which usually necessitates additional training for adaptation.  For the layer dimension, LCKV~\citep{wu2024layercondensedkvcacheefficient}, CLA~\citep{brandon2024reducingtransformerkeyvaluecache} and MLKV~\citep{zuhri2024mlkvmultilayerkeyvalueheads} reduce memory usage by sharing the KV cache across adjacent layers. ShortGPT~\citep{men2024shortgpt} and DynamicSlicing~\citep{dumitru2024change} achieve compression by eliminating redundant layers. YOCO~\citep{Sun2024YouOC} changes the model structure and shares a single global KV cache across layers. For the head dimension, MQA~\citep{shazeer2019fasttransformerdecodingwritehead} and GQA~\citep{ainslie2023gqa} share the KV cache within each head groups. DeepSeek-v2~\citep{liu2024deepseek} employs dimension-reduction techniques to compress all heads into a single low-rank vector. These lines of work are orthogonal to ours, and theoretically, they could be combined with our method. However, it is important to note that these methods typically necessitate adjustments at the model architecture or require additional fine-tuning, while quantized pruning can be directly employed without any fine-tuning.

\section{Methods}

The decoder-only transformer model consists of a stack of transformer decoder blocks, each comprising two main components: self-attention module and the feed-forward network (FFN) module.
During inference, KV cache is implemented within the self-attention module and operates in two distinct phases: i) the prefill phase, where the input prompt is used to generate KV cache for each transformer layer of LLMs; and ii) the decoding phase, where the model uses KV cache to generate the next token, and updates the KV cache with the new token. 

\paragraph{Prefill Phase.} Let $\bm{X} \in \mathbb{R}^{b \times l_{\text{prompt}} \times d}$ be the input tensor, where b is the batch size, $l_{prompt}$ is the length of the input prompt, and d is the model hidden size. For clarity, we omit the layer index here. The key and value tensors can be computed by:
\begin{align}
   \bm{X_K} = \bm{XW_K} , \bm{X_V} = \bm{XW_V} 
   \label{qkv}
\end{align}
where $\bm{W_K},\bm{W_V} \in \mathbb{R}^{d \times d}$ are the key and value layer weight. $\bm{X_K},\bm{X_V}$ are cached in the memory for utilization in the subsequent decoding phase. 

\paragraph{Decoding Phase.} Let $\bm{h} \in \mathbb{R}^{b \times 1 \times d}$ be hidden state of current input token. $\bm{h_K}=\bm{hW_K}$ and $\bm{h_V}=\bm{hW_V}$ are the current key and value states. $\bm{h_K}$ and $\bm{h_V}$ are first employed to update the KV cache:
\begin{align}
    \bm{X_K} &\xleftarrow{} \text{Concat} (\bm{X_K,h_K}), \notag \\
    \bm{X_V} &\xleftarrow{} \text{Concat} (\bm{X_V,h_V}) \label{update}
\end{align}
then attention output $\bm{h_O}$ is calculated by: 
\begin{align}
    \bm{h_O} = \text{Softmax}(\bm{h_QX_k^{T}})\bm{X_V}
    \label{attno}   
\end{align}
where $\bm{h_Q}=\bm{hW_Q}$ is the output of the query layer. For ease of illustration, we ignore the FFN module and other parts of the inference workflow that are not addressed in our approach.

\paragraph{KV Quantization}
The B-bit KV quantization process during the prefill phase can be expressed as follows: First, determine the minimum number \(z_i\) and the maximum number \(m_i\) in  $\bm{G}_i$, where $\bm{G}_i$ is a group of number in $\bm{X}_K$ or $\bm{X}_V$. Using these numbers, compute the quantized result Q($\bm{G}_i$) for each group according to the formula:

\begin{align}
Q(\bm{G}_i) &= \left\lfloor \frac{\bm{G}_i - z_i}{s_i} \right\rceil, \quad & s_i = \frac{m_i - z_i}{2^B - 1}
\end{align}
The notation \(\left\lfloor \cdot \right\rceil\) represents rounding to the nearest integer. The results from all groups are aggregated to obtain  Q($\bm{X}_K$) and Q($\bm{X}_V$).
During the decoding phase, the quantized Q($\bm{X_K}$) and Q($\bm{X_V}$) and the stored quantization parameters \(z_i\) and \(s_i\) are used to recover the original values. In the decoding phase, the dequantized result  $\bm{X_K'},\bm{X_V'}$ are used to calculate the attention output. $\bm{X_K'},\bm{X_V'}$ are obtained through aggregated $\bm{G_i'}$ for each $\bm{G_i}$. $\bm{G_i'}$ can be computed using:

\begin{equation}
\bm{G_i}' = Q(\bm{G_i}) \cdot s_x + z_x
\end{equation}

\begin{table*}[t]
\centering
\resizebox{\linewidth}{!}{
\begin{tabular}{c|cccc|cccc|cccc}
\toprule
\multirow{3}{*}{Pruning Method} & \multicolumn{12}{c}{\textit{\textbf{LongBench}}} \\
\cmidrule(l){2-5} \cmidrule(l){6-9}  \cmidrule(l){10-13}
 & \multicolumn{4}{c|}{Token=128} & \multicolumn{4}{c|}{Token=512} & \multicolumn{4}{c}{Token=2048} \\
 \cmidrule(l){2-5} \cmidrule(l){6-9}  \cmidrule(l){10-13}
 & 16-bit & 8-bit & 4-bit & 2-bit & 16-bit & 8-bit & 4-bit & 2-bit & 16-bit & 8-bit & 4-bit & 2-bit \\
 \midrule
StreamingLLM & 32.1 & 32.2 & 31.7 & \textbf{19.1} & 34.6 & 34.5 & 33.9 & \underline{20.7} & 38.1 & 38.2 & 37.8 & \textbf{23.8} \\
H2O & 35.6 & 35.6 & 34.7 & 15.8 & 37.5 & 37.4 & 36.7 & 17.7 & 39.8 & 39.7 & 39.0 & 21.1 \\
SnapKV & 35.7 & 35.7 & 35.1 & 16.6 & 40.3 & 40.4 & \underline{39.7} & 20.2 & 41.7 & 41.7 & \underline{41.0} & 22.9 \\
PyramidKV & 37.4 & 37.3 & 36.4 & \underline{17.5} & 40.3 & 40.3 & 39.6 & \textbf{20.9} & \underline{41.8} & \underline{41.8} & \textbf{41.3} & \underline{23.6} \\
Ada-KV & \underline{39.3} & \underline{39.2} & \underline{37.4} & 11.9 & \underline{40.9} & \underline{40.8} & 39.0 & 12.5 & 41.7 & 41.7 & 40.0 & 13.7 \\
HeadKV & \textbf{40.9} & \textbf{40.9} &\textbf{ 39.5} & 11.6 & \textbf{41.9} & \textbf{41.9} & \textbf{40.3} & 12.1 & \textbf{42.4} & \textbf{42.4} & 41.0 & 13.4 \\
\midrule
\multirow{3}{*}{Pruning Method} & \multicolumn{12}{c}{\textit{\textbf{Needle-in-a-Haystack}}} \\
\cmidrule(l){2-5} \cmidrule(l){6-9}  \cmidrule(l){10-13}
 & \multicolumn{4}{c|}{Token=128} & \multicolumn{4}{c|}{Token=512} & \multicolumn{4}{c}{Token=2048} \\
 \cmidrule(l){2-5} \cmidrule(l){6-9}  \cmidrule(l){10-13}
 & 16-bit & 8-bit & 4-bit & 2-bit & 16-bit & 8-bit & 4-bit & 2-bit & 16-bit & 8-bit & 4-bit & 2-bit \\
 \midrule
StreamingLLM & 27.7 & 27.7 & 27.5 & 30.9 & 35.3 & 35.3 & 35.5 & 37.3 & 66.4 & 66.5 & 66.4 & 61.8 \\
H2O & 46.9 & 46.6 & 46.8 & 36.4 & 91.2 & 91.1 & 91.0 & 54.8 & 100 & 100 & 100 & 74.1 \\
SnapKV & 83.7 & 83.7 & 82.5 & \underline{55.9} & 97.4 & 97.4 & 97.2 & \underline{66.3} & 100 & 100 & 100 & \underline{78.1} \\
PyramidKV & \textbf{98.9} & \textbf{98.9} & \textbf{98.8} & \textbf{67.5} & \textbf{100} & \textbf{100} & \textbf{100} & \textbf{78.6} & 100 & 100 & 100 & \textbf{79.6} \\
Ada-KV & 87.7 & 88.2 & 81.0 & 11.1 & 98.6 & 98.6 & 97.4 & 11.9 & 100 & 100 & 100 & 27.3 \\
HeadKV & \underline{98.6} & \underline{98.5} & \underline{98.4} & 9.1 & \underline{100} & \underline{99.9} & \underline{99.7} & 11.5 & 100 & 100 & 100 & 27.6 \\
\bottomrule
\end{tabular}
}
\caption{Feasibility of quantized pruned tokens on LongBench and Needle-in-a-Haystack with Llama-3-8B-Instruct as backbone model. We use six different KV pruning methods to retain 128, 512 and 2048 tokens, and report the results of further quantization. KIVI is used as the default quantization method, except for Ada-KV and HeadKV, which adopt FlexGen. As they retain different tokens per head, making them difficult to be compatible with KIVI. The \textbf{best} results for each token-precision setting are in \textbf{bold}, the \underline{second best} results are \underline{underlined}.} 
\label{tab:eviction_methods}
\end{table*}

\begin{figure*}[t]
    \centering
    \includegraphics[width=\linewidth]{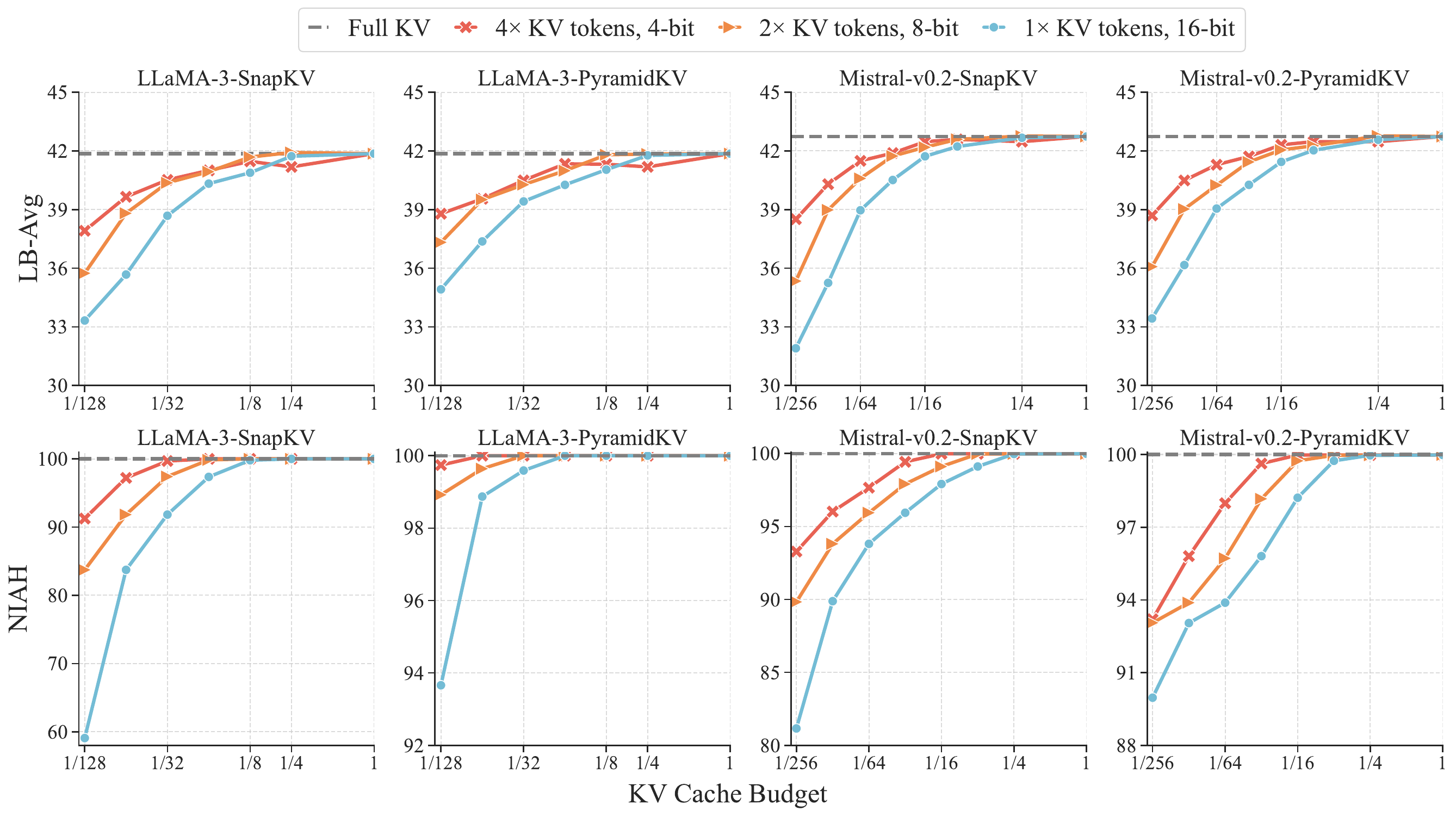}
    \caption{ The token-precision trade-off under varying memory budgets on LongBench and NIAH. We report the results of SnapKV-based and PyramidKV-based quantized pruning on Llama-3 and Mistral-v0.2. We compare three configurations with approximately equivalent memory usage: 1) Using standalone KV pruning to retain $1\times$ tokens in 16-bit precision. 2) Quantized pruning by retaining $2\times$ tokens in 8-bit precision. 3) Quantized pruning by retaining $4\times$ tokens in 4-bit precision. Quantized pruning, which preserves more tokens at a lower precision, consistently outperforms standalone KV pruning methods across various budgets.
    }
    \label{fig:budget}
\end{figure*}

\paragraph{KV Pruning}
The goal of KV pruning is to find two submatrices 
\(\bm{X}_K^e,\bm{X}_V^e \in \mathbb{R}^{b \times s \times d}\) from the full matrices \(\bm{X}_K\) and \(\bm{X}_V\) during the prefill phase, given a cache budget \(s < n\), while maximizing performance preservation. 
During the decoding phase, LLMs with KV pruning only use \(\bm{X}_K^e\) and \(\bm{X}_V^e\) to update KV cache and generate new tokens.
\begin{align}
    \bm{X}_K^e &\xleftarrow{} \text{Concat} (\bm{X_K^e,h_K}), \notag \\
    \bm{X}_V^e &\xleftarrow{} \text{Concat} (\bm{X_V^e,h_V}) 
\end{align}

\paragraph{Quantized Pruning}
Quantized pruning  uses KV pruning methods to obtain \(\bm{X}_K^e\) and \(\bm{X}_V^e\) first, and then quantizes the preserved KV states \(\bm{X}_K^e\) and 
\(\bm{X}_V^e\) to Q(\(\bm{X}_K^e\)) and Q(\(\bm{X}_V^e\)) using KV quantization methods in the prefill phase. In the decoding phase, the dequantized results from Q(\(\bm{X}_K^e\)) and Q(\(\bm{X}_V^e\)) are used to generate new tokens. 

\section{Experimental Setup}

\paragraph{Benchmarks}

We evaluate the performance of quantized pruning using the LongBench benchmark~\citep{bai-etal-2024-longbench} and Needle-in-a-Haystack test~\citep{needleinhaystack}. To better distinguish the performance impacts related to input lengths and layer-wise configurations (Sections~\ref{ssec:general} and~\ref{ssec:layer}), we further utilize RULER~\citep{hsieh2024ruler}, a dataset with different input length and diverse types of needles across 4 task categories. Detailed information on these datasets can be found in Appendix~\ref{append:dataset}.

\paragraph{LLMs}

In our primary evaluations, We employ state-of-the-art open-weight LLMs, specifically Llama-3-8B-Instruct~\citep{dubey2024llama} and Mistral-7B-Instruct-v0.2~\citep{jiang2023mistral7b}. 
For scaling experiments in Section~\ref{ssec:scale}, we additionally assess the performance of Llama-3.2-1B, Llama-3.2-3B~\citep{dubey2024llama} and Llama-3-70B~\citep{dubey2024llama}.

\paragraph{Setup}
Our experiments are designed to comprehensively investigate the token-precision trade-off in KV cache compression. We quantify memory budget by reporting the compression ratio of the KV cache relative to the full, uncompressed KV cache. For KV pruning, we utilize PyramidKV~\citep{cai2024pyramidkvdynamickvcache} and SnapKV~\citep{li2024snapkvllmknowslooking} as representative state-of-the-art methods.  To assess the feasibility of quantized pruning (Section~\ref{ssec:main}), we also include StreamingLLM~\citep{xiao2023streamingllm}, H2O~\citep{10.5555/3666122.3667628}, Ada-KV~\citep{feng2024adakvoptimizingkvcache}, and HeadKV~\citep{fu2024not}. For KV quantization, we adopt KIVI~\citep{pmlr-v235-liu24bz} as our primary method due to its stability and broad compatibility.  Furthermore, in Section~\ref{ssec:quant_strategy}, we evaluate quantization strategies from FlexGen~\citep{sheng2023flexgenhighthroughputgenerativeinference} and KVQuant~\citep{hooper2024kvquant10millioncontext} for a comprehensive comparative analysis. Additional setup details are provided in Appendix~\ref{append:setup}.

\section{Optimal Token-Precision Trade-Off}
\label{ssec:main}

In this section, we aim to find the optimal token-precision trade-off in KV cache compression. We first examine the feasibility of integrating KV pruning and quantization(Q1). Subsequently, we explore the best optimal allocation strategy between precision and token under varying memory budgets(Q2).

\finding{
Q1. Is it feasible to integrate KV pruning and quantization for a lower compression rate?
}

We first evaluate the feasibility of integrating KV pruning and quantization as a prerequisite for exploring the token-precision trade-off.
We employ Llama-3-8B-Instruct and evaluate a range of KV pruning methods on the LongBench and NIAH. We report the results of quantizing the remaining tokens to different precision levels after applying KV pruning.


From Table~\ref{tab:eviction_methods}, we observe that \textit{it is feasible to quantize pruned KV cache for a lower compression rate.}
For most KV pruning methods we evaluate, further quantizing the preserved tokens to as low as 4-bit precision results in minimal performance degradation, while quantizing to 8-bit precision shows negligible impact.  However, reducing precision to 2-bit leads to a drastic performance decline across most KV pruning methods.
This observation holds consistently across different KV pruning methods and varying numbers of preserved tokens. 

Notably, head-level KV allocation methods such as Ada-KV and HeadKV, which are incompatible with KIVI, perform well under high precision and a low number of preserved tokens. However, when precision is reduced to as low as 2-bit, these methods experience a more pronounced performance degradation compared to KV pruning methods compatible with KIVI, such as SnapKV and PyramidKV. This observation underscores the critical importance of considering compatibility with KV quantization techniques in the development of future KV pruning strategies.

Compared with precision, reducing the number of preserved tokens leads to more significant performance degradation.
Specifically, when the number of preserved tokens is reduced to 1/4 (from 2048 to 512), all KV pruning methods experience a noticeable performance drop. 
In contrast, when the precision is reduced to 1/4 (from 16-bit to 4-bit), which has the same memory budget  as token dimension, the performance degradation is relatively mild. 
This suggests that, under the same memory budget, tokens might have a more significant impact on the results compared to precision. 

To conclude, KV pruning can be effectively integrated with KV quantization at a precision level of 4-bit without substantial performance degradation.  


\finding{
Q2. What is the optimal allocation strategy between precision and token  under varying memory budgets?
}

Observing that KV pruning and KV quantization can be effectively integrated, we further investigate that, given a fixed memory budget, how to balance the trade-off between number of preserved tokens and precision to achieve optimal performance.
To this end, we evaluate the performance of quantized pruning using two leading KV pruning methods, SnapKV and PyramidKV, across different memory budgets on LongBench and NIAH.


As shown in Figure~\ref{fig:budget}, we observe that \textit{quantized pruning, which preserves more tokens at a lower precision, consistently outperforms standalone KV pruning methods across various budgets.} 
For the NIAH task, the improvements from quantized pruning are particularly pronounced.
This may be attributed to that quantized pruning can cover more tokens for retrieval under the same memory budget compared to standalone KV pruning.

In high-budget scenarios, the 8-bit strategy tends to deliver slightly better performance, which may due to the number of tokens at this budget is already quite large.
In low-budget scenarios, such as 1/128 KV cache budget, storing more tokens at 4-bit precision yields superior results, highlighting the importance of token coverage when resources are constrained.
Overall, using lower precision to preserve more tokens under a limited budget results in notable performance gains, compared to standalone KV pruning methods that use full precision to store fewer tokens.

\paragraph{Summary}
We demonstrate that storing \textbf{\textit{more tokens}} in the KV cache with \textbf{\textit{lower precision}} can significantly enhance the long-context performance of LLMs under fixed KV cache memory budget.

\section{Further Analysis}

In this section, we further investigate series of key aspects regarding token-precision trade-off, including the impact of quantized pruning on various downstream task types and input lengths, model scaling effect, ablation on quantization strategies and  fine-grained exploration of layer-wise quantized pruning.

\begin{table*}[t]
\centering
\resizebox{\linewidth}{!}{
\begin{tabular}{c|cc|ccccccc}
\toprule
\multirow{2}{*}{Models} & \multirow{2}{*}{Token} & \multirow{2}{*}{Bit} & \multicolumn{7}{c}{Task Types} \\
\cmidrule(l){4-10}
 &  &  & SQA & MQA & SUMM & Fewshot & Syn. & Code & RULER-8k \\
 \midrule
\multirow{3}{*}{Llama-3-8B-Instruct} & 512 & 16 & 28.2 & 31.9 & 23.5 & 67.6 & \textbf{37.7} & 57.6 & 67.5 \\
 & 1024 & 8 & 29.6 & \textbf{33.1} & 24.3 & 67.9 & 37.4 &\textbf{ 58.0} & 74.9 \\
 & 2048 & 4 & \textbf{30.7} & 32.5 & \textbf{25.3} &\textbf{ 68.8} & 37.2 & 57.6 & \textbf{82.2} \\
 \midrule
\multirow{3}{*}{Mistral-7B-Instruct-v0.2} & 512 & 16 & 33.7 & 27.3 & 24.3 & 65.6 & 41.75 & 54.0 & 53.1 \\
 & 1024 & 8 & 34.2 & \textbf{29.0} & 25.6 & 66.4 & \textbf{43.73} & 54.8 & 62.1 \\
 & 2048 & 4 & \textbf{35.2} & 28.14 & \textbf{26.6} & \textbf{66.9} & 43.08 & \textbf{55.4} & \textbf{73.6} \\
 \bottomrule
\end{tabular}
}
\caption{The token-precision trade-off in different task types. We report the results of 6 task types in LongBench and 8k subset of RULER. We use PyramidKV-based quantized pruning. The \textbf{best} results for each task type are in \textbf{bold}.}
\label{tab:tasks}
\end{table*}

\subsection{Impact on Task Types and Input Lengths }
\label{ssec:general}

\paragraph{
Task Types
}

To further investigate the token-precision trade-off in different task types, we evaluate PyramidKV-based quantized pruning on six task types from LongBench and the 8K subset of the RULER dataset.
We use 1/16 KV cache budget, and explore the token-precision trade-off under this fixed memory budget, as this setting exhibits minimal performance differences across three settings, facilitating a more direct comparison of performance across task types.

As illustrated in Table~\ref{tab:tasks}, we observe that the performance of quantized pruning is remarkably consistent across different task types. Specifically, lower precision, which retains more tokens in KV cache, leads to substantial performance improvements in the RULER task, which heavily relies on retrieving contents from the input. 
Tasks with high retrieval demands, such as Summarization and Single-Doc QA, also show noticeable gains with quantized pruning, particularly when $4\times$ tokens are preserved at 4-bit precision. 

For tasks requiring more reasoning rather than intensive retrieval, such as Code Completion, Synthetic and Multi-Doc QA, the benefits of trading precision for more tokens are less pronounced. 
In these cases, storing fewer tokens with higher precision generally performs better. For example, using 1024 tokens in 8-bit precision achieves the hightest score of 58 in Code task.


\begin{figure}[t]
    \centering
    \includegraphics[width=\linewidth]{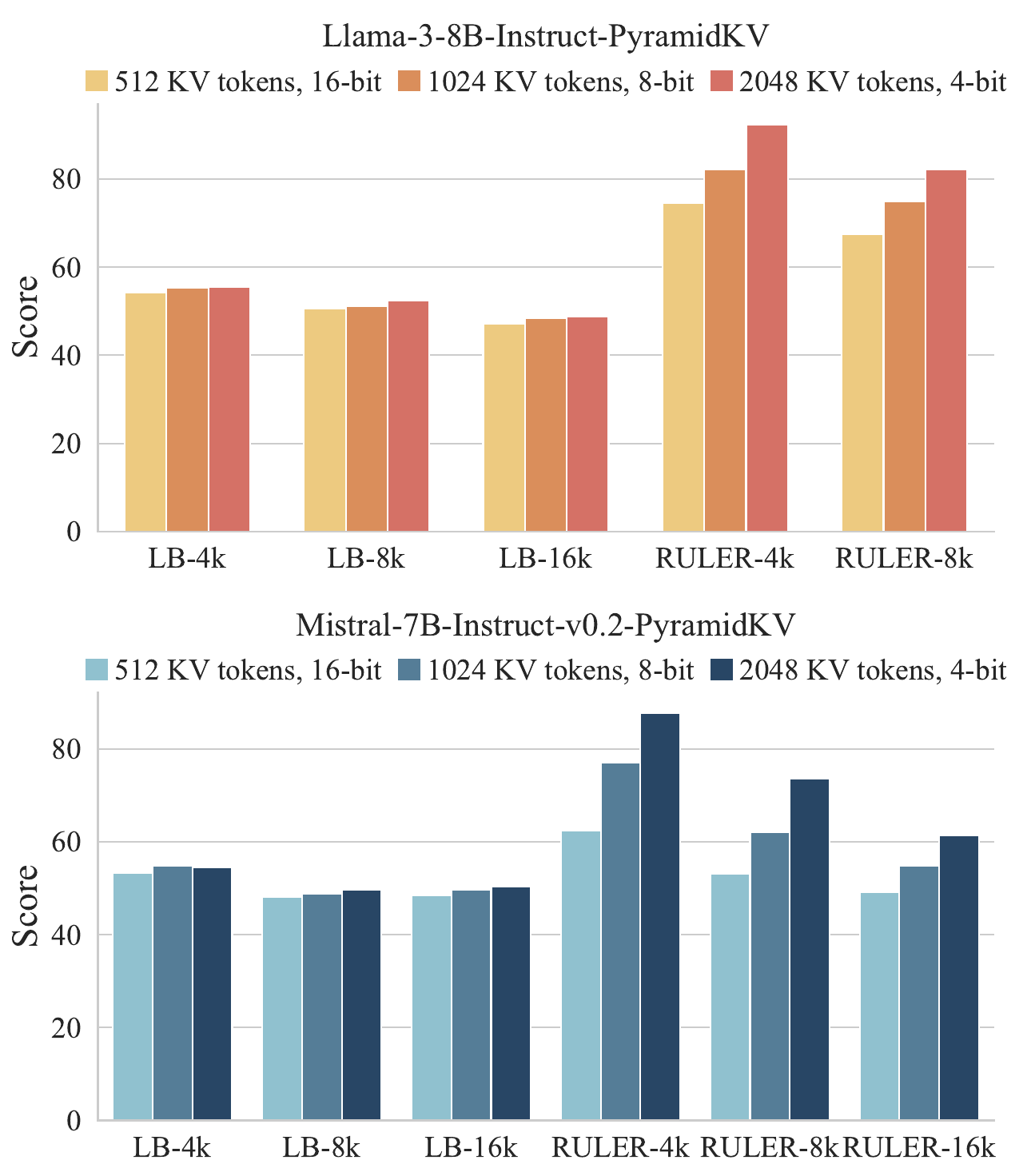}
    \caption{
    The token-precision trade-off in different input lengths. We report the results of LongBench and three subsets of RULER. We use PyramidKV-based quantized pruning.
    }
    \label{fig:intput_length}
\end{figure}

\paragraph{Input Lengths}
To evaluate the token-precision trade-off across various input lengths, we conduct experiments on subsets with different input length of the RULER dataset.
Additionally, we analyze LongBench by grouping its data based on input length, more detailed information can be found in Appendix~\ref{append:dataset}.
The results are shown in Figure~\ref{fig:intput_length}.

Our observations are as follows: \textit{quantized pruning consistently outperforms standalone KV eviction methods across various input lengths, regardless of the models and task types.}
Within the same dataset, scores decrease as input length increases; however, the relative differences among different compression methods remain similar across varying input lengths.
Moreover, quantized pruning achieves significant performance improvements across all input lengths for retrieval demanded tasks like RULER.

\subsection{Scaling Effect on Quantized Pruning}
\label{ssec:scale}

\begin{figure}[t]
    \centering
    \includegraphics[width=\linewidth]{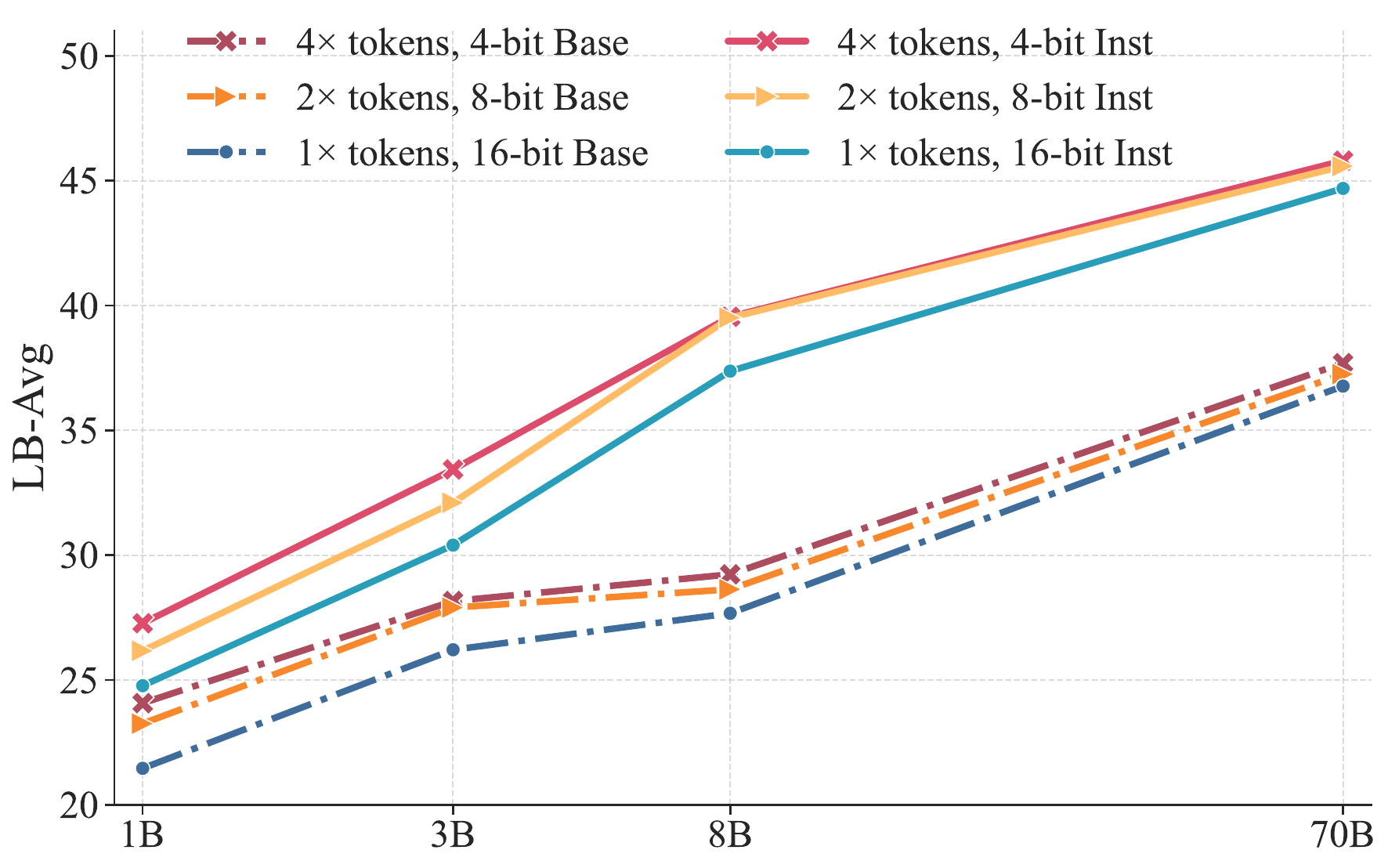}
    \caption{
    Scaling effect on Llama family models, with
    PyramidKV-based quantized pruning. All models are under 1/64 KV cache budget. 
    }
    \label{fig:scale}
\end{figure}

To investigate the impact of model scaling on quantized pruning, we conducted experiments on four models from the Llama series: Llama3-70B, Llama3-8B, Llama3.2-3B and Llama3.2-1B. Since the Llama3 series does not include 1B and 3B versions, we used the Llama3.2 series as substitutes. However, it is important to note that the performance of Llama3.2-3B-Base is quite similar to Llama3-8B-Base. For both the Base 
 models and Instruct models, we evaluated their performance on LongBench under 1/64 KV cache budgets. To further validate the conclusion, we also experimented with a 1/16 KV cache budget, and the results are presented in the Appendix~\ref{para:scale}. 

As shown in Figure~\ref{fig:scale}, we observe that quantized pruning consistently achieves better performance across all scaling levels. Notably, the performance gap between quantized pruning and standalone KV pruning methods remains relatively stable across different model scales. Interestingly, as the model scale increases, the performance gap between 2× tokens with 8-bit precision and 4× tokens with 4-bit precision becomes smaller. This may be because deeper models encode more redundant information within each token, allowing a relatively smaller number of tokens to retain sufficient information.  
For Base models, although the performance improvement from scaling is smaller compared to Instruct models, quantized pruning still provides a noticeable performance boost. 

These findings highlighting the robustness and effectiveness of quantized pruning across model scaling.

\subsection{Ablation on Quantization Strategies}
\label{ssec:quant_strategy}

While there has been extensive research on strategies for KV quantization, it remains unclear whether existing quantization strategies remain effective when integrated with KV pruning methods. In this section, we aim to investigate the impact of KV quantization strategies on quantized pruning. We also conducted an analysis of the effect of quantization group size on the quantized pruning, with further results available in Appendix~\ref{para:group}.

\begin{figure}[t]
    \centering
    \includegraphics[width=\linewidth]{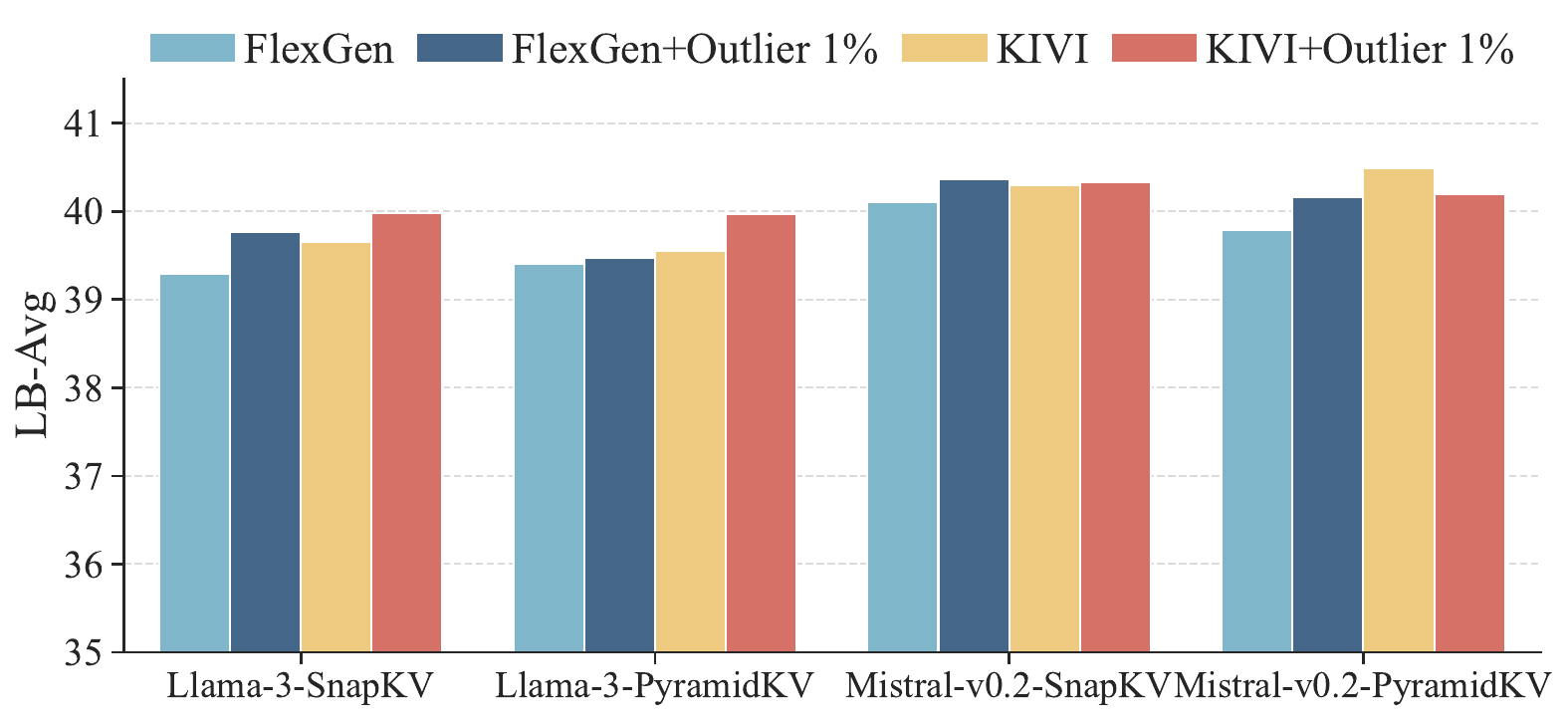}
    \caption{
    Ablation of quantization strategies on quantized pruning, remaining 512 KV tokens in 4-bit. 
    }
    \label{fig:quant}
\end{figure}

\paragraph{Quantization methods} We explore the methods in FlexGen~\citep{sheng2023flexgenhighthroughputgenerativeinference}, KIVI~\citep{pmlr-v235-liu24bz}, and KVQuant~\citep{hooper2024kvquant10millioncontext}. To elaborate, for the FlexGen methods, KV quantization is applied to both the key and value caches along the token dimension, grouping every 64 elements without filtering outlier numbers. We modify the FlexGen by (1) filtering 1\% of outlier numbers in both the key and value cache, as mentioned in KVQuant (2) quantizing the key along the channel dimension, as in KIVI and (3) combining (1) and (2). These correspond to the results labeled as FlexGen+Outlier 1\%, KIVI, and KIVI+Outlier 1\% in the Figure~\ref{fig:quant}.

We can observe that none of the quantization strategies show significant performance degradation when combined with KV pruning methods, demonstrating the relative stability of quantized pruning.
The KIVI method consistently outperforms FlexGen across various models and KV pruning methods. The improvement is particularly pronounced for PyramidKV on the Mistral model, underscoring the significance of quantizing key states along the channel dimension.
Filtering 1\% of outlier numbers proves effective for the FlexGen strategy but yields limited improvements for KIVI. It shows some benefit on Llama models but results in negligible gains on the Mistral models.

Overall, KIVI demonstrates strong performance when integrated with KV pruning methods, while other KV quantization strategies
also maintain good results, highlighting the stability of quantized pruning.

\subsection{Exploration on Layer-Wise Quantized Pruning}
\label{ssec:layer}

\begin{figure}[t]
    \centering
    \includegraphics[width=1.0\linewidth]{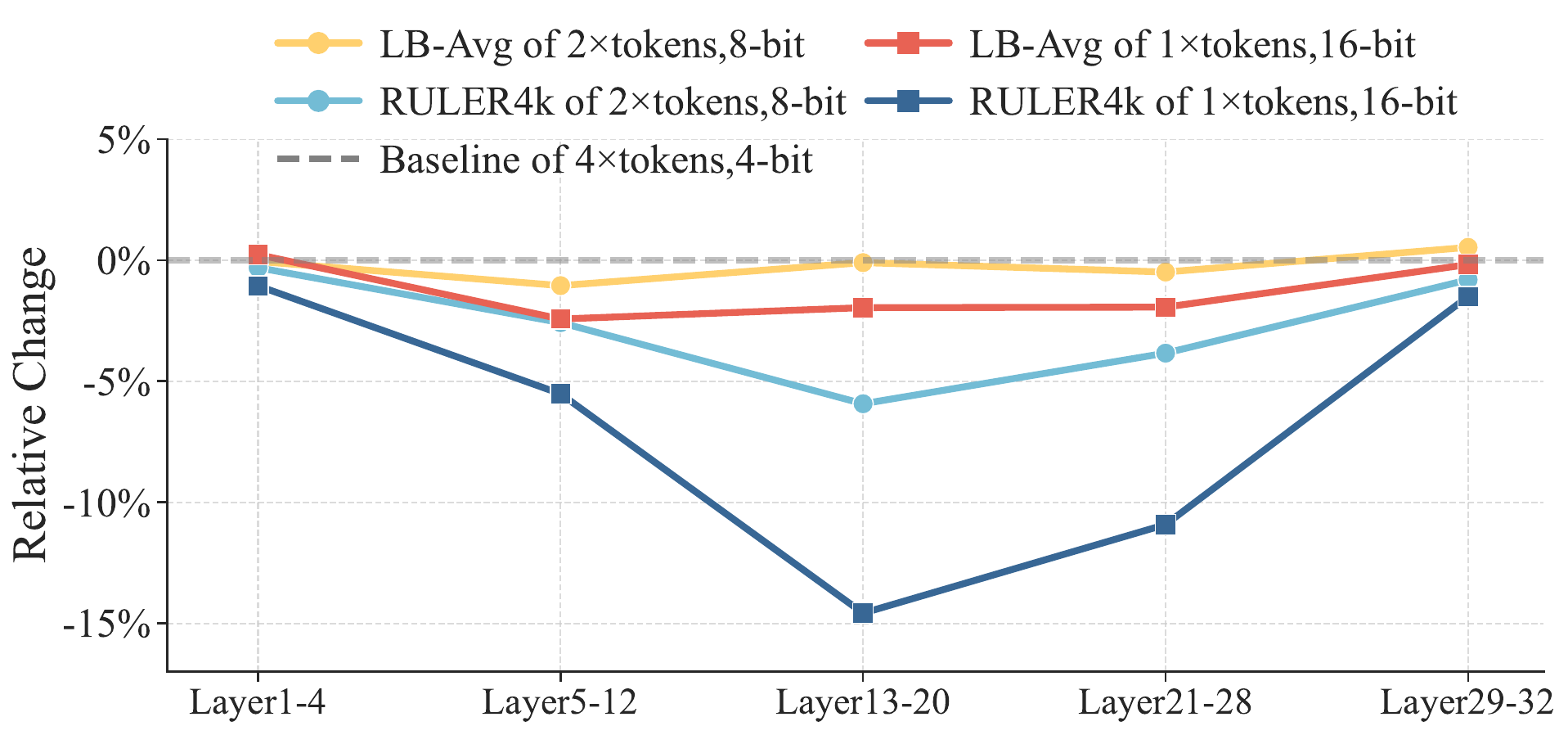}
    \caption{
    The results of layer-wise quantized pruning on Llama-3-8B-Instruct, with SnapKV as pruning method. We use $4\times$ KV token 4-bit as baseline and report the relative change. Configurations are modified every 4 layers for the initial and final layers, while intermediate layers are reconfigured every 8 layers.  
    }
    \label{fig:layer}
\end{figure}

Inspired by the observation that different layers may have varying requirements for the number of tokens in PyramidKV~\citep{cai2024pyramidkvdynamickvcache} and PyramidInfer~\citep{yang2024pyramidinferpyramidkvcache}, we further investigate whether the demands for precision and preserved tokens are consistent across layers. To explore this, we use the best-performing configuration from previous experiments, 4-bit precision with $4\times$ tokens, as the baseline and compare it against layer-wise configurations adopting 8-bit precision with $2\times$ tokens and 16-bit precision with $1\times$ tokens. Using SnapKV as the KV pruning method, we present the results for Llama-3 under 1/64 KV cache budget in the Figure~\ref{fig:layer}. The x-axis shows the modified layers range, while the y-axis shows the relative change to the baseline (4-bit precision with $4\times$ tokens) on LongBench and RULER-4k. Furthermore, we present the results for Llama-3 evaluated with a 1/16 KV cache budget and Mistral-v0.2 evaluated with 1/64 and 1/16 KV cache budgets. These detailed results are available in the Appendix~\ref{para:layer}.

It is evident that for most layers, transitioning from $4\times$ tokens with 4-bit precision to higher precision and fewer tokens results in a performance decline under constrained KV cache budgets.
Specifically, the shift to 8-bit shows a relatively minor performance drop, whereas moving to 16-bit with fewer preserved tokens leads to a more significant decrease. 
These layers-wise trade-off conclusions are consistent with our experiments before.

Notably, modifying intermediate layers causes a drastic performance decline, while adjustments made at the initial and final layers result in comparatively smaller performance reductions. This effect is especially pronounced in retrieval-related tasks such as RULER-4k, where significant performance differences are observed. On LongBench, changes are less significant, with a notable performance drop only observed at 16-bit precision. These findings highlight that, under the same memory budget, \textit{preserving more tokens with lower precision in intermediate layers is crucial for the performance}, while the token-precision trade-off  in the initial and final layers exerts a more balanced influence.

\section{Conclusion}

We investigate a series of critical yet unexplored questions regarding the effectiveness and feasibility of token-precision trade-off in KV cache compression. Through comprehensive experiments, we demonstrate that storing more tokens in the KV cache with lower precision can significantly enhance the long-context performance of LLMs, and demonstrating robust performance across diverse input lengths, downstream tasks, with particularly significant gains in retrieval tasks. Moreover, we show that quantized pruning demonstrates strong feasibility across different KV pruning methods, quantization strategies, and model scales.
Our analysis sheds light on the token-precision trade-off of KV cache memory optimization, offering valuable insights into designing more efficient compression strategies. We hope this work deepens the understanding of KV  cache compression and inspires future research.

\section*{Limitations}

While our work demonstrates the effectiveness of KV compression through trade-offs between token and precision dimensions, other potential dimensions, such as head and layer, remain unexplored. Investigating the feasibility of combining these dimensions with token and precision for a more substantial compression potential represents an avenue for future research. Additionally, the current implementation of quantized pruning suffers from inefficiencies in dequantizing the KV cache, hindering the full realization of speedup benefits from the memory savings.  In future work, we aim to address this issue by optimizing the implementation, such as integrating fusion operators to combine dequantization with matrix multiplication.
\bibliography{custom}
\appendix

\section{Datasets}
\label{append:dataset}

\paragraph{LongBench} 
LongBench~\citep{bai-etal-2024-longbench} includes 17 datasets covering 6 categories of tasks, which can be divided into single-document QA~\citep{dasigi2021dataset,kovcisky2018narrativeqa}, multi-document QA~\citep{yang2018hotpotqa,ho2020constructing}, summarization~\citep{huang2021efficient,fabbri2019multi,zhong2021qmsum}, few-shot learning~\citep{gliwa2019samsum,joshi2017triviaqa,li2002learning}, synthetic, and code generation~\citep{guo2023longcoder,liu2023repobench}. LongBench features an average input length ranging from 1,235 to 18,409 tokens. For inputs exceeding the model's context window length(8k for Llama-3-8B-Instruct~\citep{dubey2024llama}, we split the data and only take the beginning and end segments of the input to  fill the context window length. Additionally, we reserve sufficient space for newly generated tokens based on the specific type of sub-dataset. For evaluate the impact of input lengths, we select datasets with sufficient data to cover three input length ranges: (<4k, 4k~8k, and >8k). These datasets include MultiFieldQA-en, 2WikiMultihopQA, GovReport, TREC, TriviaQA, SAMSum, and RepoBench-P, representing a variety of task types. We refer to the three subsets as LB-4k, LB-8k, and LB-16k, respectively.

\paragraph{NIAH}
 Needle-in-a-Haystack(NIAH)~\citep{needleinhaystack} is a challenging pressure test designed to assess the ability of models to accurate identify and retrieve relevant information from lengthy context.  NIAH randomly inserts key information into an arbitrary position within a long essay. In our setup, we use PaulGrahamEssays as the haystack and the sentence "The best thing to do in San Francisco is eat a sandwich and sit in Dolores Park on a sunny day." as the needle, which is the default setting of NIAH. We vary the essay length from 1,000 tokens up to the models' context window limits, increasing by 100 tokens per step for Llama-series models and 400 tokens per step for Mistral. The results are reported as the average score across all tests.

\begin{figure}[t]
    \centering
    \includegraphics[width=1.0\linewidth]{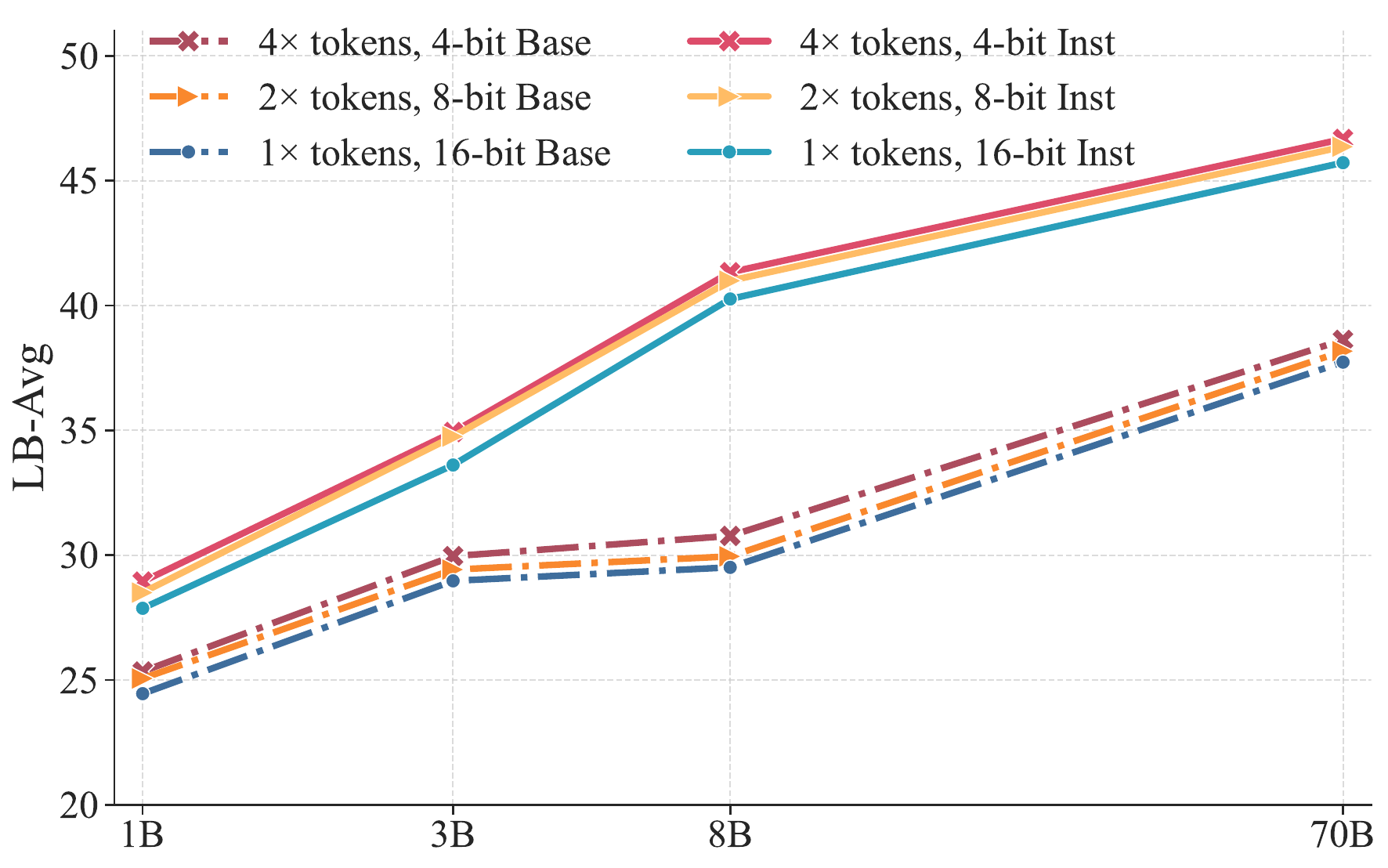}
    \caption{
    Scaling effect on Llama family models, with
PyramidKV-based quantized pruning. All models are
under 1/16 KV cache budget.
    }
    \label{fig:scale16}
\end{figure}

\paragraph{RULER}

RULER~\citep{hsieh2024ruler} generates synthetic examples to evaluate long-context language models with configurable sequence lengths and varying task complexities. It includes four task categories: Retrieval, Multi-hop Tracing, Aggregation, and Question Answering. The dataset comprises six subsets with input lengths of 4K, 8K, 16K, 32K, 64K and 128K tokens. In our experiments, we use the 4K, 8K and 16K subsets to test the models within their context window limits.

\section{Experiment Setup}
\label{append:setup}
\paragraph{Memory Budgets}  We report the ratio of compressed KV cache and the full KV cache for memory budge. The full KV cache for Llama-3 is 8k KV tokens in 16-bit on both LongBench and NIAH, while for Mistral-v0.2 is 16k KV tokens on LongBench and 32k KV tokens on NIAH in 16-bit.
\paragraph{KV eviction methods} We retain the last 32 tokens for StreamingLLM~\citep{xiao2023streamingllm}, H2O~\citep{10.5555/3666122.3667628}, and SnapKV~\citep{li2024snapkvllmknowslooking}, while keeping 8 tokens for PyramidKV~\citep{cai2024pyramidkvdynamickvcache}, Ada-KV~\citep{feng2024adakvoptimizingkvcache} and HeadKV~\citep{fu2024not}, as recommended in the corresponding paper~\citep{cai2024pyramidkvdynamickvcache,fu2024not}. For other settings, we adopt the default configurations from their papers.

\paragraph{KV quantization} We utilize  HQQQuantizedCache from Huggingface and adjust the group dimensions of keys and values to implement grouped quantization strategies from FlexGen~\citep{sheng2023flexgenhighthroughputgenerativeinference} and KIVI~\citep{pmlr-v235-liu24bz}. We use 64 as the default group size which is suggested in FlexGen~\citep{sheng2023flexgenhighthroughputgenerativeinference}. In the experiments involving outlier filtering, we exclude numbers in the KV cache with a absolute value exceeding 6 from quantization, which roughly corresponds to the top 1\% of outliers based on our validation set analysis.

\section{More results in Experiments}
\paragraph{Group Size} 
\label{para:group}
We analyze the impact of group size during KV quantization. We employ SnapKV and PyramidKV to retain 512 tokens with 4-bit KIVI quantization and report the performance variations when the group sizes were set to 32, 64 and 128. As shown in Table~\ref{tab:group}, smaller group sizes lead to performance improvements at the cost of higher memory usage. Reducing the group size from 128 to 64 resulted in a notable improvement, but further decreasing it from 64 to 32 yielded minimal gains for the Mistral model. Therefore, we set the default quantization group size to 64 to balance performance and memory usage in our experiments.
\begin{table}[t]
\centering
\resizebox{\linewidth}{!}{%
\begin{tabular}{ccccc}
\toprule
\multirow{2.5}{*}{Model} & \multirow{2.5}{*}{Method} & \multicolumn{3}{c}{Group Size} \\ \cmidrule(r){3-5}
                       &                         & 32       & 64      & 128      \\  
\midrule
\multirow{2}{*}{Llama-3} & SnapKV & 40.4 & 39.6 & 38.9 \\
 & PyramidKV & 40.3 & 39.6 & 38.9 \\
 \midrule
\multirow{2}{*}{Mistral-v0.2} & SnapKV & 40.4 & 40.3 & 40.1 \\
 & PyramidKV & 40.3 & 40.5 & 40.0 \\
\bottomrule
\end{tabular}%
}
\caption{The impact of group size for quantized pruning on LongBench. }
\label{tab:group}
\end{table}

\paragraph{Scaling Effect on Quantized Pruning}
\label{para:scale}

We validate our findings by measuring performance at a budget of 1/16, with the results shown in Figure~\ref{fig:scale16}. This corroborates the conclusion reported in the main text using a budget of 1/64.  When the KV cache budget is relative small to 1/64 , the performance improvement brought by quantized pruning is higher compared to 1/16 KV cache budget, which aligns with the conclusions we observed earlier in Q2.

\begin{figure}[t]
    \centering
    \includegraphics[width=1.0\linewidth]{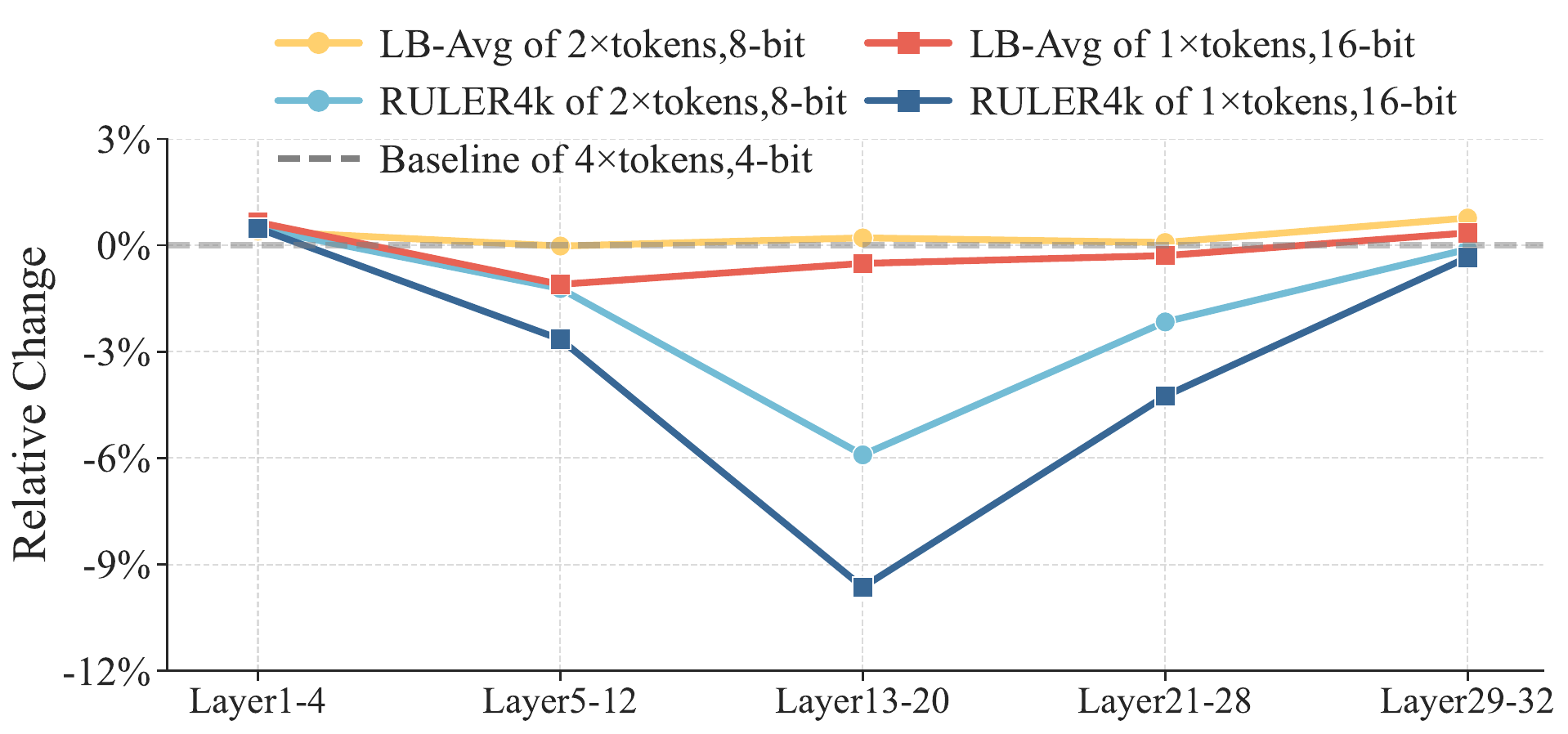}
    \caption{
    The results of Layer-Wise Quantized Pruning on Llama-3-8B-Instruct, with SnapKV as pruning method. KV cache budget=1/16.
    }
    \label{fig:layer16}
\end{figure}

\begin{figure}[t]
    \centering
    \includegraphics[width=1.0\linewidth]{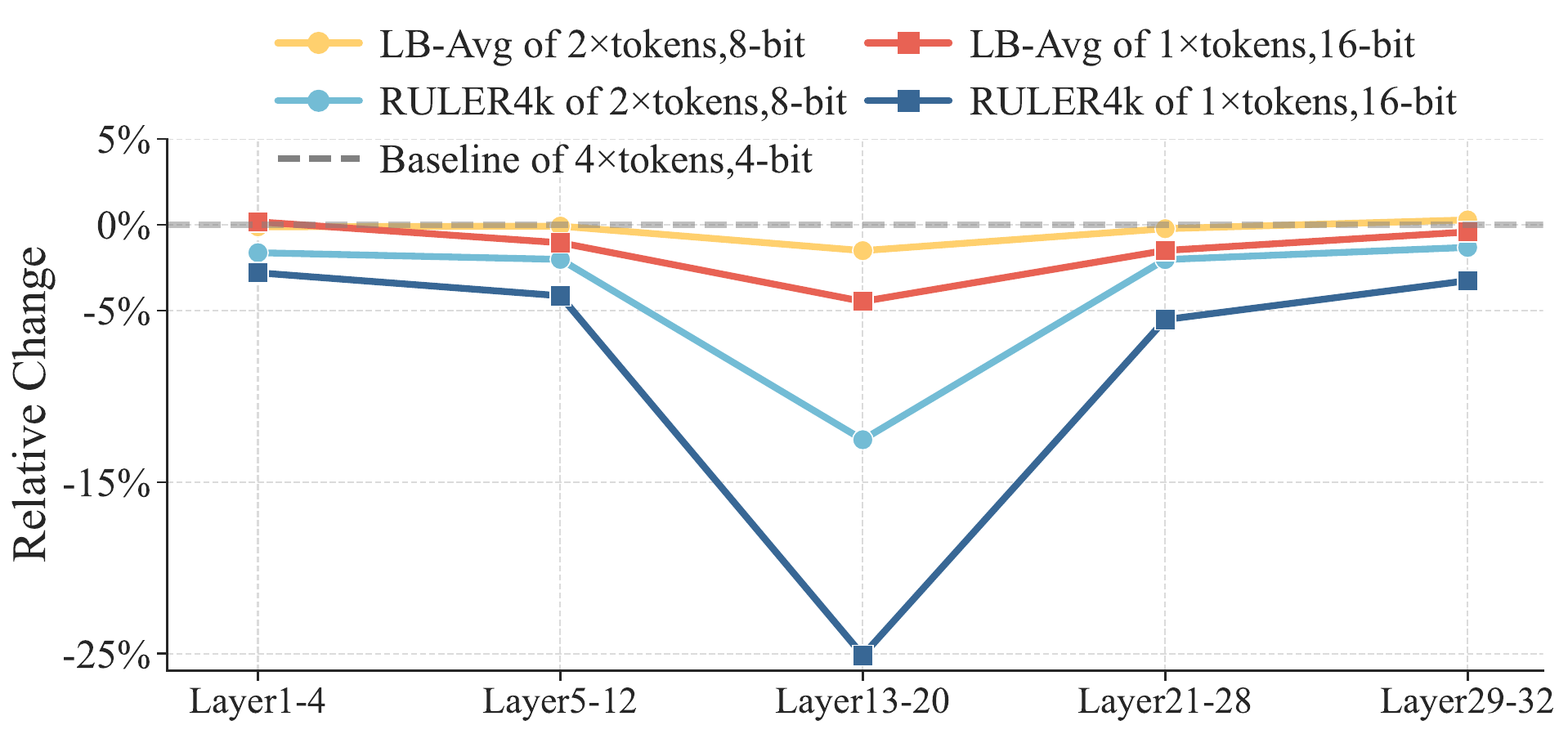}
    \caption{
    The results of Layer-Wise Quantized Pruning on Mistral-7B-v0.2-Instruct, with SnapKV as pruning method. KV cache budget=1/64.
    }
    \label{fig:layermistral}
\end{figure}

\begin{figure}[htbp]
    \centering
    \includegraphics[width=1.0\linewidth]{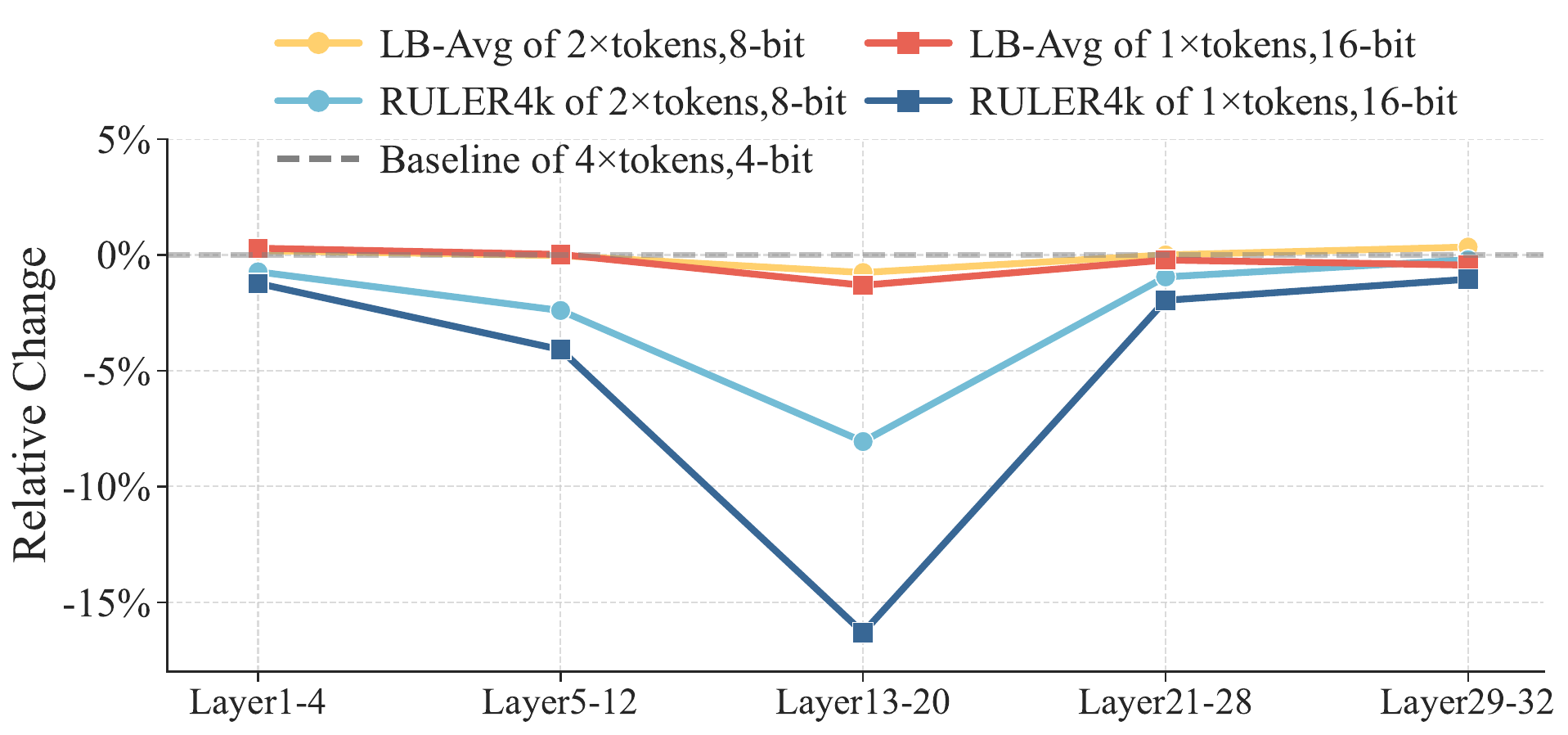}
    \caption{
    The results of Layer-Wise Quantized Pruning on Mistral-7B-v0.2-Instruct, with SnapKV as pruning method. KV cache budget=1/16.
    }
    \label{fig:layermistral16}
\end{figure}

\paragraph{Layer-Wise Quantized Pruning}
\label{para:layer}

We further validated our findings by measuring performance of Llama-3 at a budget of 1/16, with the results shown in Figure~\ref{fig:layer16}.
We alse report the results for Mistral-v0.2 in Figure~\ref{fig:layermistral} and Figure~\ref{fig:layermistral16}, we can see the layer-wise results are similiar to Llama-3. Modifying intermediate layers causes a drastic performance decline, while adjustments made at the initial and final layers result in comparatively smaller performance reductions. This effect is especially pronounced in retrieval-related tasks such as RULER-4k, where significant performance differences are observed. On LongBench, changes are less significant, with a notable performance drop only observed at 16-bit precision.

\end{document}